\documentclass[10pt, a4paper]{article}

\usepackage[]{lrec-coling2024} 

\usepackage{makecell}

\usepackage{color}
\usepackage{caption}
\usepackage{subcaption}
\usepackage{threeparttable}

\usepackage{diagbox}
\usepackage{booktabs}

\title{
DuetSim: Building User Simulator with Dual Large Language Models for Task-Oriented Dialogues
}

\name{Xiang Luo, Zhiwen Tang{*}\thanks{*Corresponding Authors}, Jin Wang{*} and Xuejie Zhang}

\address{School of Information Science and Engineering \\
         Yunnan University \\
         Kunming, China\\
        luoxiang@mail.ynu.edu.cn, \{zhiwen.tang, wangjin, xjzhang\}@ynu.edu.cn\\}

\abstract{
User Simulators play a pivotal role in training and evaluating task-oriented dialogue systems. Traditional user simulators typically rely on human-engineered agendas, resulting in generated responses that often lack diversity and spontaneity. Although large language models (LLMs) exhibit a remarkable capacity for generating coherent and contextually appropriate utterances, they may fall short when tasked with generating responses that effectively guide users towards their goals, particularly in dialogues with intricate constraints and requirements.
This paper introduces DuetSim, a novel framework designed to address the intricate demands of task-oriented dialogues by leveraging LLMs. DuetSim stands apart from conventional approaches by employing two LLMs in tandem: one dedicated to response generation and the other focused on verification. This dual LLM approach empowers DuetSim to produce responses that not only exhibit diversity but also demonstrate accuracy and are preferred by human users. We validate the efficacy of our method through extensive experiments conducted on the MultiWOZ dataset, highlighting improvements in response quality and correctness, largely attributed to the incorporation of the second LLM.
Our code is accessible at: \url{https://github.com/suntea233/DuetSim}.
 \\ \newline  }

\begin{document}

\maketitleabstract

\section{Introduction}


\begin{figure*}[t]
    \centering
    \begin{subfigure}{\textwidth}
        \includegraphics[width=0.95\textwidth]{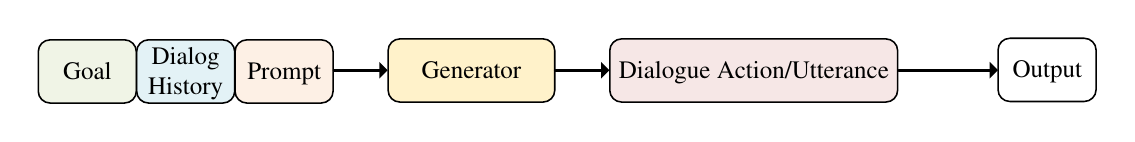}
        \caption{User Simulator with a single LLM for generator}
        \label{fig:single_llm}

    \end{subfigure}

    \begin{subfigure}{\textwidth}
        \includegraphics[width=0.95\textwidth]{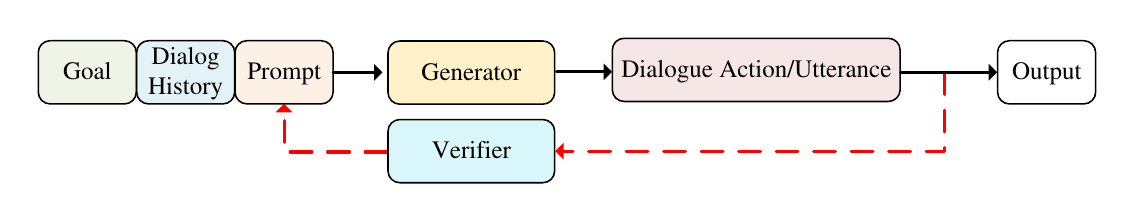}

        \caption{User Simulator with dual LLMs for generator and verifier}
        \label{fig:dual_llm}
    \end{subfigure}
    \caption{Different Architectures for LLM-based User Simulators}
    \label{fig:different generative approaches}
\end{figure*}

Task-oriented dialogue systems (TODS) are developed to engage with human users in natural language to accomplish tasks within specific domains, such as restaurant reservations and ticket bookings. A common approach to training TODS involves reinforcement learning, where the dialogue system engages with human users over tens of thousands of interactions to collect training dialogues, optimize dialogue policies, and enhance generated responses. However, operating within this training paradigm, which relies on interactions with actual human users, is labor-intensive and cost-prohibitive. To address this challenge, user simulators (US) have been introduced to emulate the behavior of real human users, thereby substituting them in the reinforcement learning training process for TODS.

User simulators play a crucial role in the training of TODS. They not only reduce training costs but also significantly influence the quality of the system's responses, thereby impacting the overall performance of TODS. Given this importance, researchers have long focused on developing user simulators capable of producing accurate, diverse, and human-like responses. Early efforts \cite{schatzmann-etal-2007-agenda,schatztnann2005effects} centered on constructing user simulators using expert knowledge and handcrafted rules. However, rule-based user simulators face several challenges. One challenge is the complexity of human engineering required, which restricts their suitability to small, well-defined domains and makes them less adaptable for extension to new domains. Another challenge is the limited diversity in their responses, stemming from the determinism of the rules that guide them and the predefined templates used for generating natural language utterances. Although rule-based user simulators can produce correct responses most of the time, researchers have explored alternative approaches to address the limitations of rule-based simulators.

The advent of deep learning (DL) has also given rise to DL-based user simulators. Notable examples of such work include \cite{asri2016sequence,user-izzeddin-2018,kreyssig-etal-2018-neural,lin-etal-2021-domain,lin-etal-2022-gentus}. DL-based user simulators offer significant improvements in language variation, but they require extensive training on large volumes of human-annotated task-oriented dialogue data. The quality and diversity of this training data play a pivotal role in determining the accuracy and robustness of these user simulators, which also renders them less suitable for transfer to new domains, particularly when resources are limited in the new domain.

The advent of large language models (LLMs) and in-context learning has opened up new possibilities for constructing user simulators. LLMs are trained on vast amounts of text data in a self-supervised manner and exhibit impressive zero-shot and few-shot capabilities in downstream tasks, often requiring only a small number of examples to perform well. An intriguing direction is to utilize LLMs as user simulators by providing them with appropriate prompts. This approach circumvents the need for training and fine-tuning, making it more adaptable and faster in accommodating new domains. However, such endeavors \cite{terragni2023context} are not universally successful. Given the intricate requirements of task-oriented dialogues, LLM-based user simulators struggle to consistently generate responses that satisfy all the criteria specified in the prompts. This study reveals that the challenge arises from LLMs' limitations in comprehending lengthy prompts. Our findings align with those of \cite{liu2023lost}, which highlight that LLMs are more inclined to focus on information at the beginning or end of input, with model performance diminishing as input length increases.

In order to address the aforementioned challenges, this paper introduces a novel framework, DuetSim, designed to tackle the complex requirements of task-oriented dialogues using LLMs. Instead of relying on a single LLM to construct the user simulator, we advocate the use of two LLMs, one for the generator and the other for the verifier, to build a more robust user simulator. The generator initially generates responses based on the dialogue context, while the verifier meticulously examines these responses, offering feedback to the generator if any generated responses are deemed unsuitable. Figure \ref{fig:different generative approaches} highlights the distinctions between our approach and those employing a single LLM. With the inclusion of the verifier, DuetSim has the capacity to rectify erroneous responses identified by the verifier.
Additionally, we incorporate chain-of-thought reasoning to narrow the search space, aiding the user simulator in comprehending the dialogue context and generating contextually appropriate dialogue responses.

We conducted experiments with our approach using the MultiWOZ dataset. The experimental results clearly demonstrate that our proposed method generates responses that exhibit greater diversity, accuracy, and are preferred by human users. The incorporation of the second LLM significantly enhances the quality and correctness of the generated responses.

The remainder of this paper is structured as follows: Section 2 provides a brief review of related work. In Section 3, we offer a comprehensive description of our model. Section 4 provides a summary of the experimental results and implementation details. Finally, in Section 5, we draw our conclusions.

\section{Related Work}
\subsection{User Simulator}
Task-oriented dialogue system is designed to interact with users in a goal-directed manner to accomplish specific tasks or provide relevant information \cite{wen-etal-2017-network,ham-etal-2020-end}. They differ from chit-chat dialogue systems, which are more focused on free-form conversations and entertaining interactions, while task-oriented dialogue systems are geared towards addressing the specific needs of users. 

User simulators also play a crucial role in task-oriented dialog systems. Different from dialogue systems, user simulators have a goal generator to guide them in simulating user behavior and evaluate the performance of the user simulator based on the extent to which the goals are achieved. Typically, constructing a user simulator can be achieved through an agenda-based approach \cite{schatztnann2005effects,schatzmann2006survey,schatzmann-etal-2007-agenda,keizer-etal-2010-parameter}, where rules are designed based on expert knowledge to simulate user behavior. Additionally, there are deep learning-based methods that involve training models on corpora of dialogues, resulting in models that don't require hand-crafted policy and have good scalability \cite{asri2016sequence,user-izzeddin-2018,kreyssig-etal-2018-neural,lin-etal-2021-domain,lin-etal-2022-gentus}. 
Different from the methods mentioned above, we directly use large language models to generate dialogue actions and utterances. Through this approach, we do not need to train a model from scratch, and it offers excellent transferability.


\subsection{Large Language Models and Prompt Learning}
In recent years, with the emergence of GPT-3 \cite{brown2020language}, large language models have achieved remarkable performance across various tasks \cite{zhao2023survey,qiao-etal-2023-reasoning,yuan2022hierarchical}. Additionally, the few-shot and even zero-shot learning capabilities demonstrated by these models have been particularly impressive. As research on prompt learning has evolved, we've seen a progression from initially guiding large language models through simple prompts for downstream tasks to more advanced techniques, which includes step-by-step guided reasoning using the chain of thought approach \cite{chain-wei-2022}, as well as recent developments involving strategies like tree-structured architectures \cite{yao2023tree}. These approaches aim to effectively apply large language models to various tasks using different decoding methods \cite{zheng2023progressive}. Recently, there have also been many efforts to use prompt learning to guide large language models in exploring various downstream tasks such as dialogue system \cite{hudecek-dusek-2023-large}, dialog state tracking \cite{wang-etal-2022-slot}, user simulator \cite{terragni2023context}. As done by Terragni et al. \cite{terragni2023context}, sampling a few examples from the dataset helps large language models better adapt to playing user simulator.

With the deepening research into hard prompts, there has been exploration of approaches like training additional verifier to verify the correctness of the final answer \cite{cobbe2021training} or intermediate steps \cite{li-etal-2023-making,lightman2023let}. Besides, there are methods that involve verifying generated code to improve result accuracy \cite{zhou2023solving}. In contrast to these approaches, which either rely on a single large language model to perform all tasks or involve training an additional model to provide feedback, we utilize two large language models, solely harnessing their inference capabilities and facilitating interaction between the generator and the verifier to enhance the applicability of the entire model in user simulation.


\section{DuetSim}

\begin{figure*}[!t]
    \centering
    \includegraphics[width=0.95\textwidth]{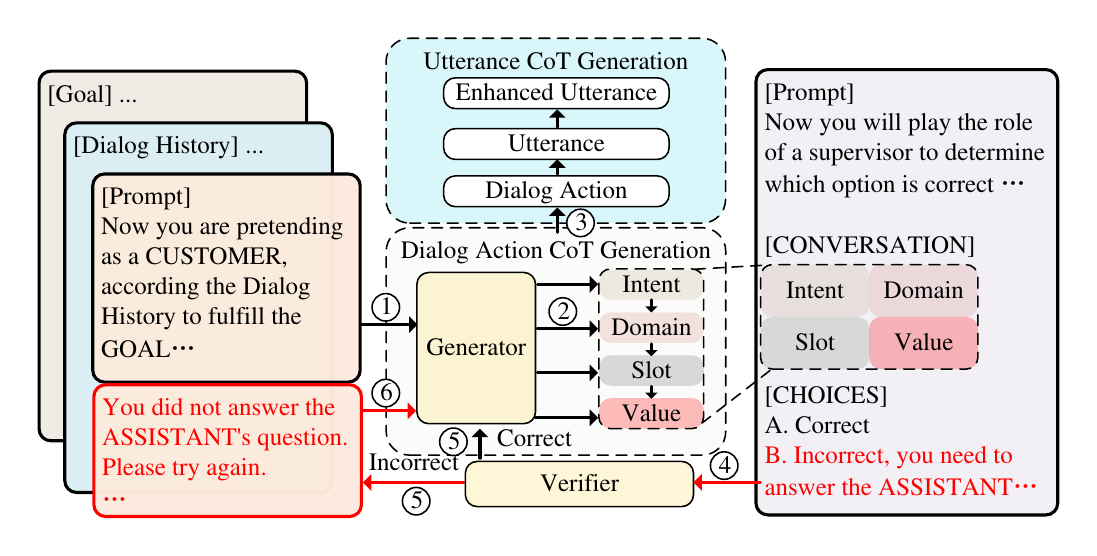}
    \caption{Details of DuetSim.}
    \label{fig:interaction}
\end{figure*}

In this paper, we propose DuetSim, a user simulator based on two LLMs for task-oriented dialogue systems. DuetSim consists of a dialogue generator and a response verifier. The dialogue generator drafts the response while the response verifier examines the generated response and offers feedback if needed.
Both the dialogue generator and the response verifier are LLMs. By splitting the task between the dialogue generator and the response verifier, both LLMs can share the burden and better handle the assigned tasks. Further, we also propose a chain-of-thought approach to guide the user simulator to generate responses that better fit the context. Rather than directly generating responses in natural language, the proposed approach first generates dialogue acts step-by-step, which are then used to guide the utterance generation. 

\subsection{Prompt Learning for DuetSim}

DuetSim leverages the capabilities of large language models (LLMs) with prompt learning. We create prompts for both the dialogue generator and response verifier to elicit responses from LLMs. By supplying LLMs with background information from the ongoing dialogue and previous conversation history, both models can effectively perform their designated tasks. This includes generating appropriate responses in dialogue acts or natural language and verifying whether the generated responses meet the specified requirements.

We employ a zero-shot learning approach to harness the inference capabilities of LLMs. In this approach, we abstain from providing any demonstrations but instead list the requirements and prompt the LLM to generate a response. The prompt for the dialogue generator includes essential information such as the target user goal, denoted as $G$, the dialogue context, represented as $C$, and the requirements, denoted as $R_c$. The user goal $G$ is randomly generated, and the entire dialogue is designed to fulfill this user goal. The dialogue context $C$ encompasses utterances from previous dialogue turns, while the requirements ${R^G} = [R_1^G, R_2^G, ..., R_n^G]$ specify the desired properties of the generated response.
The task of the dialogue generator is to produce appropriate dialogue acts. Each dialogue act comprises four components: intent, domain, slot, and value, serving as semantic abstractions for the dialogue utterances. This process can be formulated as follows:

\begin{equation}\label{generator_output}
    {U^G} = Generator\left( {G,C,{R^G}} \right)
\end{equation}

\noindent where the ${U^G}$ represent the output for dialogue generator. 
The left side of Figure \ref{fig:interaction} shows the prompt learning process for the dialogue generator. 




The output from the dialogue generator is then passed to the response verifier to assess its alignment with the context. The prompt for the response verifier includes the requirements denoted as $R_v$, the dialogue context represented as $C$, and the response generated by the dialogue generator, denoted as $U^G$. Within the requirement set ${R^V} = [R_1^V, R_2^V, ..., R_m^V]$, we instruct the verifier LLM to scrutinize the response ${U^G}$ for any potential errors. In these requirements, we enumerate common mistakes that the LLM might make, such as context inconsistency and the generation of meaningless words. This prompts the verifier to check for any such errors in the generated response. This process can be formulated as follows:

\begin{equation}\label{eqaution:verifier output}
    {U^V} = Verifier\left( {G,C,{R^V},{U^G}} \right)
\end{equation}

where the ${U^V}$ represent the output for verifier.
The right side of Figure \ref{fig:interaction} shows the prompt learning process for the response verifier.


\subsection{Generator-Verifier Interaction}
In DuetSim, instead of relying on a single LLM for generating dialogue responses, we introduce an additional response verifier to enhance the quality of the generated responses. The dialogue generator initially produces a draft response based on provided prompts. Subsequently, this draft response is forwarded to the response verifier for evaluation against specified requirements. If the verifier deems the draft generated response as suitable, it becomes the output of the user simulator. Conversely, if the verifier detects any errors in the draft response, it rejects the response and provides feedback to the generator. The generator then proceeds to generate an alternative response.


Figure \ref{fig:interaction} illustrates the interactive process. In the case of the generator, this process entails inputting carefully crafted prompts into a large language model and leveraging the model's reasoning capabilities to generate dialogue acts through a chain of thought. Following this, the generated dialogue acts, along with their context, are then transmitted to the verifier.

As for the verifier, it combines the received dialogue acts with pre-defined prompts and inputs them into a large language model. Likewise, the verifier utilizes the reasoning capabilities of the large language model to evaluate the accuracy of the dialogue acts. Subsequently, the verifier offers feedback to the dialogue generator regarding the correctness of the generated response.






If the draft response is found to be incorrect, the feedback will encompass the specific requirement $R^V_i$ that the draft response has breached. The generator subsequently integrates this feedback information at the end of the prompt and proceeds to generate another response.
This iterative process persists until the generator successfully generates the correct dialogue actions or reaches the maximum specified number of iterations. As a result, the final output is formulated as:

\begin{equation}
    \widetilde {{U^G}} = Generator\left( {G,C,{R^G},{U^V}} \right)
\end{equation}

\begin{table}[!t]
\resizebox{0.49\textwidth}{!}{
\begin{tabular}{ll}
\hline
Procedure                                & Example                              \\ \hline
\makecell{Dialog Action\\-\textgreater{}Utterance}    & \begin{tabular}[c]{@{}l@{}}{[}EXAMPLE{]}\\ {[}{[}'inform', 'restaurant', 'book day', 'Tuesday'{]}{]}\\ The restaurant is booked on Tuesday.\\ {[}END EXAMPLE{]}\\  \{Dialog Action\}\\ Please translate the list into natural language.\end{tabular}              \\ \hline

\makecell{Utterance-\textgreater{}\\Enhanced Utterance} & \begin{tabular}[c]{@{}l@{}}{[}CONVERSATION{]}\\ \{CONVERSATION\} \\ {[}SENTENCE{]}\\ \{Utterance\} \\ Based on conversation, play the role of \\ CUSTOMER and rewrite this sentence to make \\ it smoother, more natural, and more conversational\end{tabular} \\ \hline
\end{tabular}
}
\caption{Response Generation with CoT}
\label{table: example of prompt}
\end{table}

\begin{table*}[ht]
\resizebox{0.98\textwidth}{!}{
\begin{tabular}{llllllll}
\hline
User Simulator                        & Complete Rate & Success Rate & Precision & Recall & $F_1$-score & Book Rate & Turn  \\ \hline
ABUS                            & \textbf{0.97}          & \textbf{0.97}         & \textbf{0.902}     & \textbf{0.983}  & \textbf{0.924}    & \textbf{0.970}     & 10.36 \\
PBUS                            & 0.41          & 0.30         & 0.580     & 0.670  & 0.710    & 0.659     & \textbf{7.50}\footnotemark[2]   \\ \hline
DuetSim (ChatGLM2)               & 0.19          & 0.25         & 0.699     & 0.646  & 0.644    & 0.332     & 17.22 \\
DuetSim (LLAMA2)                 & 0.19          & 0.28         & 0.748     & 0.688  & 0.687    & 0.060     & 17.42 \\ \hline
DuetSim (ChatGPT)                & 0.92          & 0.74         & 0.830     & 0.980  & 0.881    & 0.585     & 16.92 \\
DuetSim (ChatGPT w/o verifier) & 0.85          & 0.67         & 0.842     & 0.948  & 0.873    & 0.544     & 17.88 \\
DuetSim (FLAN-T5)                & 0.92          & 0.71         & 0.820     & 0.979  & 0.872    & 0.648     & 16.16 \\
DuetSim (FLAN-T5 w/o verifier) & 0.90          & 0.63         & 0.834     & 0.961  & 0.875    & 0.551     & 16.18 \\ \hline
\end{tabular}
}
\caption{Goal Fulfillment Comparison on MultiWOZ dataset}
\label{table:semantic-level}
\end{table*}
\footnotetext[2]{PBUS enforces a maximum 15-turn limit for dialogues, unlike other models.}

\subsection{Chain-of-thought Response Generation}
To enhance natural language generation, we use a "chain of thought" approach. Initially, we provide guidance through examples, enabling the large language model to effectively translate dialogue actions. As a foundation, we boost fluency, contextual relevance, and naturalness in generated utterances by integrating dialogue context and prompts. This approach yields higher-quality output compared to direct generation. We start by converting dialogue actions into simple utterances using examples and crafted prompts. Subsequently, we input this information, along with the dialogue context, into the large language model, leveraging its reasoning capabilities to produce coherent, context-aware utterances. See Table \ref{table: example of prompt} for an illustrative example prompt.

To enhance dialogue action generation, we employ a step-by-step "chain of thought" approach. Directly generating dialogue actions, including $intent, domain, slot, and \enspace value$, can be challenging for large language models due to their limited understanding of abstract dialogue actions. This makes it difficult to produce well-structured actions aligned with intended goals. Thus, we adopt a method that breaks down the generation process into sequential steps for each component of dialogue actions. This approach reduces complexity, assisting the user simulator in producing precise dialogue actions. Specifically, we begin by generating the $intent$ using a dedicated prompt. Once obtained, this $intent$ serves as input for the $domain$ prompt. We continue this sequential process for $slot$ and $value$. This iterative approach allows us to construct complete dialogue actions by gradually generating each component.

\section{Experiments}


\begin{table*}[]
\resizebox{0.98\textwidth}{!}{
\begin{tabular}{lllllllll}
\hline
User Simulator         & Unigrams      & Bigrams       & Trigrams      & Entropy       & CE           & MSTTR         & HDD          & MTLD           \\ \hline
ABUS-T            & 446           & 1641          & 2717          & 6.87          & 2.39         & 0.71          & 0.75         & 45.97          \\
ABUS-S            & 413           & 1514          & 2386          & 7.04          & 2.37         & 0.76          & 0.79         & 62.35          \\
PBUS              & 949           & 4440          & \textbf{7176} & 7.40          & \textbf{3.0} & 0.70          & 0.78         & 45.50          \\ \hline
DuetSim (ChatGPT)  & 1032          & 2781          & 3894          & 7.44          & 2.62         & \textbf{0.77} & 0.78         & 56.98          \\
DuetSim (FLAN-T5)  & 916           & 2274          & 2818          & \textbf{7.58} & 2.73         & 0.74          & 0.79         & 50.63          \\
DuetSim (ChatGLM2) & 1288          & 3913          & 5793          & 7.51          & 2.31         & 0.74          & \textbf{0.8} & 55.28          \\
DuetSim (LLAMA2)   & \textbf{1421} & \textbf{4446} & 6753          & 7.11          & 2.14         & 0.77          & 0.75         & \textbf{63.75} \\ \hline
\end{tabular}
}
\caption{Utterance Diversity Comparison on MultiWOZ dataset}
\label{table:utterance-level}
\end{table*}



\begin{table*}[ht]
\centering
\resizebox{.98\textwidth}{!}{
\begin{tabular}{llllllll}
\hline
US Model           & Complete Rate & Success Rate & Precision & Recall & $F_1$-score    & Book Rate & Turn  \\ \hline
DuetSim          & 0.92          & 0.74         & 0.830      & 0.980   & 0.881 & 0.585     & 16.92 \\
w/o Dialogue history & 0.85          & 0.68         & 0.793      & 0.904   & 0.824 & 0.626     & 17.86  \\
w/o Goal           & 0.91          & 0.70          & 0.815     & 0.980   & 0.871 & 0.455     & 16.52 \\
w/o Both           & 0.89          & 0.66          & 0.821     & 0.939   & 0.859 & 0.584     & 17.46 \\
\hline
\end{tabular}
}
\caption{Ablation results.}
\label{ablation}
\end{table*}

\begin{table*}[ht]
\resizebox{0.98\textwidth}{!}{
\begin{tabular}{lccccccc}
\hline
User Simulator                               & Complete Rate & Success Rate & Precision & Recall & $F_1$-score    & Book Rate & Turn  \\ \hline
\makecell[l]{DuetSim w. dialogue\\ context in utterance}
& \textbf{0.92}          & \textbf{0.74}         & 0.830      & \textbf{0.980}   & 0.881 & \textbf{0.585}     & 16.92 \\
\hline
\makecell[l]{DuetSim w. dialogue\\ context in dialogue act} & 0.89          & 0.66         & \textbf{0.853}     & 0.968  & \textbf{0.890} & 0.523     & \textbf{16.60} \\ \hline
\end{tabular}
}
\caption{DuetSim with different forms of dialogue context.}
\label{type of history}
\end{table*}


\subsection{Dataset}
We evaluate DuetSim using the MultiWOZ dataset \cite{budzianowski-etal-2018-multiwoz} available in ConvLab \cite{zhu-etal-2020-convlab}. MultiWOZ is an extensively annotated dataset comprising 10,000 human-to-human written conversations covering diverse domains and topics, making it a widely used benchmark dataset for evaluating task-oriented dialogue systems.


\subsection{Evaluation Metrics}
We evaluate the proposed method from two perspectives, goal fulfillment and utterance diversity. While goal fulfillment metrics evaluate whether the dialogue helps the user complete the task, utterance diversity metrics focus on the quality of generated  natural language. 

Goal fulfillment metrics include success rate, completion rate, booking rate, precision, recall and $F_1$-score. 
Success rate refers to the fraction of dialogues that successfully accomplish the user's task. 
Completion rate  evaluates whether the dialogue system make reservations disregarding whether the reserved entity match the user's requirement. 
Book rate assesses  whether booking tasks present in the user's goal have been completed. Precision, recall and $F_1$-score are used to determine whether the dialogue system finds the required information.


The utterance diversity are measured with multiple different metrics, including number of unique n-grams (unigrams, bigrams, and trigrams), Shannon Entropy (SE) \cite{manning1999foundations}, Conditional bigram Entropy (CE) \cite{manning1999foundations}, Mean Segmental Type-Token Ratio (MSTTR) \cite{lu2012relationship}, Measure of Textual Lexical Diversity (MTLD) \cite{mccarthy2010mtld} and Hypergeometric Distribution Function (HDD)  \cite{wu1993accurate}. Among them, SE measures the diversity of information contained in the text. CE assesses the fluency and coherence of text. MSTTR measures the diversity and lexical richness of text. MTLD additionally considers the overall structure of text when measuring the lexical diversity.  HDD measures the diversity of vocabulary when randomly selecting a fixed number of words from the text.

In addition, we conducted human evaluation to assess whether responses generated by DuetSim aligns with human preferences. 


\subsection{Implementation Details}
All the experiment results are the average of 100 task-oriented dialogues. All the user simulators interact with the dialogue system via dialogue acts. We implement the proposed method using different LLMs, including ChatGPT \footnote{\url{https://openai.com/chatgpt}}, FLAN-T5 \cite{chung2022scaling}, LLAMA2 \cite{touvron2023llama} and ChatGLM2 \cite{du-etal-2022-glm}.


\subsection{Baselines}

We compare the proposed method with Agenda-Based User Simulator (ABUS) \cite{schatzmann-etal-2007-agenda} and Prompt-Based User Simulator (PBUS) \cite{terragni2023context}.

\begin{itemize}
    \item \noindent {\bf ABUS}: ABUS is an user simulator that follows hand-crafted rules to complete dialogues. We adopt two variants, namely ABUS-T that uses template NLG and ABUS-S that uses SC-GPT for NLG when evaluating utterance diversity. 
    

    \item \noindent {\bf PBUS}: PBUS is based on a single LLM and uses in-context learning to simulate users. Compared with our approach, PBUS does not involve chain-of-thought reasoning when simulating users. 
    
\end{itemize}

The proposed method are respectively implemented using the following LLMs:

\begin{itemize}

    \item \noindent {\bf ChatGPT}. ChatGPT is a chatbot developed by OpenAI. It builds upon the foundation of GPT and is trained on large volumes of text data to comprehend and generate natural language text.

    \item \noindent {\bf FLAN-T5}. FLAN-T5 is an instruction-tuned model, trained on extensive datasets, that excels in almost all downstream tasks due to its high versatility.

    \item \noindent {\bf LLAMA2}. LLAMA2 is a model based on LLAMA, and it builds upon the foundation of LLAMA by improving the quality of training data, increasing context length, and optimizing memory related to caching.

    \item \noindent {\bf ChatGLM2}. ChatGLM2 is a model based on ChatGLM. Building on ChatGLM, ChatGLM2 upgrades the base model to enhance performance, increases context length, and improves inference capabilities.

\end{itemize}

\subsection{Main Results}

\textbf{Goal Fulfillment}: As shown in Table \ref{table:semantic-level}, 
DuetSim performs best in all the user simulators that are based on LLMs and the performance of DuetSim(ChatGPT) is the closest to ABUS. Among all the variants of DuetSim, 
DuetSim(ChatGPT) and DuetSim(FLAN-T5) perform much better than 
DuetSim(ChatGLM2) and DuetSim(LLAMA2), which can be attributed to the strong inference capabilities of ChatGPT and FLAN-T5.  ChatGPT's strong inference capabilities are largely due to its large parameter count, whereas for FLAN-T5, the benefits arise from its architecture, data, and instruction fine-tuning, which enable it to also possess good inference capabilities.


We further compare with variants of DuetSim that do not include response verifier. We include all the requirements in the prompt for the single LLM. Results show that performance deteriorate after removing the response verifier from the user simulator. The utterance generation with CoT and the interaction between response verifier and dialogue generator greatly improve the performance of DuetSim.


\textbf{Utterance Diversity}: Table \ref{table:utterance-level} presents the performance of different models in terms of utterance diversity using various automated metrics, which demonstrates that our utterance generation mechanism based on CoT significantly improves the language diversity in all metrics. Due to differences in the training data, architecture, and parameter counts of various models, there are variations in performance among these models as well.


\subsection{Ablation Study}
To investigate the effectiveness of different components in our prompts, we conducted an ablation experiment on both the goal and dialogue history parts. The results of ablation experiments in Table \ref{ablation} demonstrate the effectiveness of the two components in our prompts. Removing dialogue history or user goal will results in the worsening of performance since both contains important information for understanding the current status of the dialogue. 

We further investigate the impact of different forms of dialogue context, namely dialogue acts or natural language utterances. Results in Table \ref{type of history} show that DuetSim is better at comprehending dialogue context in the form of natural language, which supports our initial design that uses natural language to express dialogue history.



\subsection{Cross-model Evaluation}
An ideal user simulator should help the dialogue system obtain better generalization ability. With such observation, we conduct a cross-model evaluation where we train the dialogue system on a user simulator first, then test the dialogue system on another simulator. The training of dialogue system is driven by reinforcement learning algorithm, i.e. proximal policy optimization (PPO) \cite{schulman2017proximal}. The experiment results are reported in \ref{table:cross-model}.

We observe that the dialogue system training on DuetSim and testing on ABUS performs much better than the one training  on ABUS and testing on DuetSim. This indicates that training on DuetSim greatly improves the generalization ability of the dialogue system. 

In the meantime, we also observe that, when training and testing the dialogue system on the same user simulator, dialogue system that interacts with DuetSim perform much worse than the that interacts with ABUS. This indicates that responding to dialogues generated by DuetSim is more challenging than interacting with ABUS. 

The above differences come from the fact ABUS is driven by human-engineered agenda, which means the generated response may lack the diversity and stochasticity that are better handled by LLMs. Training the dialogue system using DuetSim is more challenging and but it also improves the generalizability for the dialogue system.


\begin{table}[!t]
\resizebox{0.49\textwidth}{!}{
\begin{tabular}{l|l|l}
\toprule
\diagbox{Training}{Testing} & DuetSim       & ABUS          \\ \midrule
{DuetSim}      & Complete 0.55 & Complete 0.87 \\
                              & Success 0.46  & Success 0.83  \\
\hline
{ABUS}         & Complete 0.46 & Complete 0.91 \\
                              & Success 0.41  & Success 0.88  \\ \bottomrule
\end{tabular}
}
\caption{Cross-model evaluation results.}
\label{table:cross-model}
\end{table}


\begin{figure*}[!t]
    \centering
    \includegraphics[width=1\textwidth]{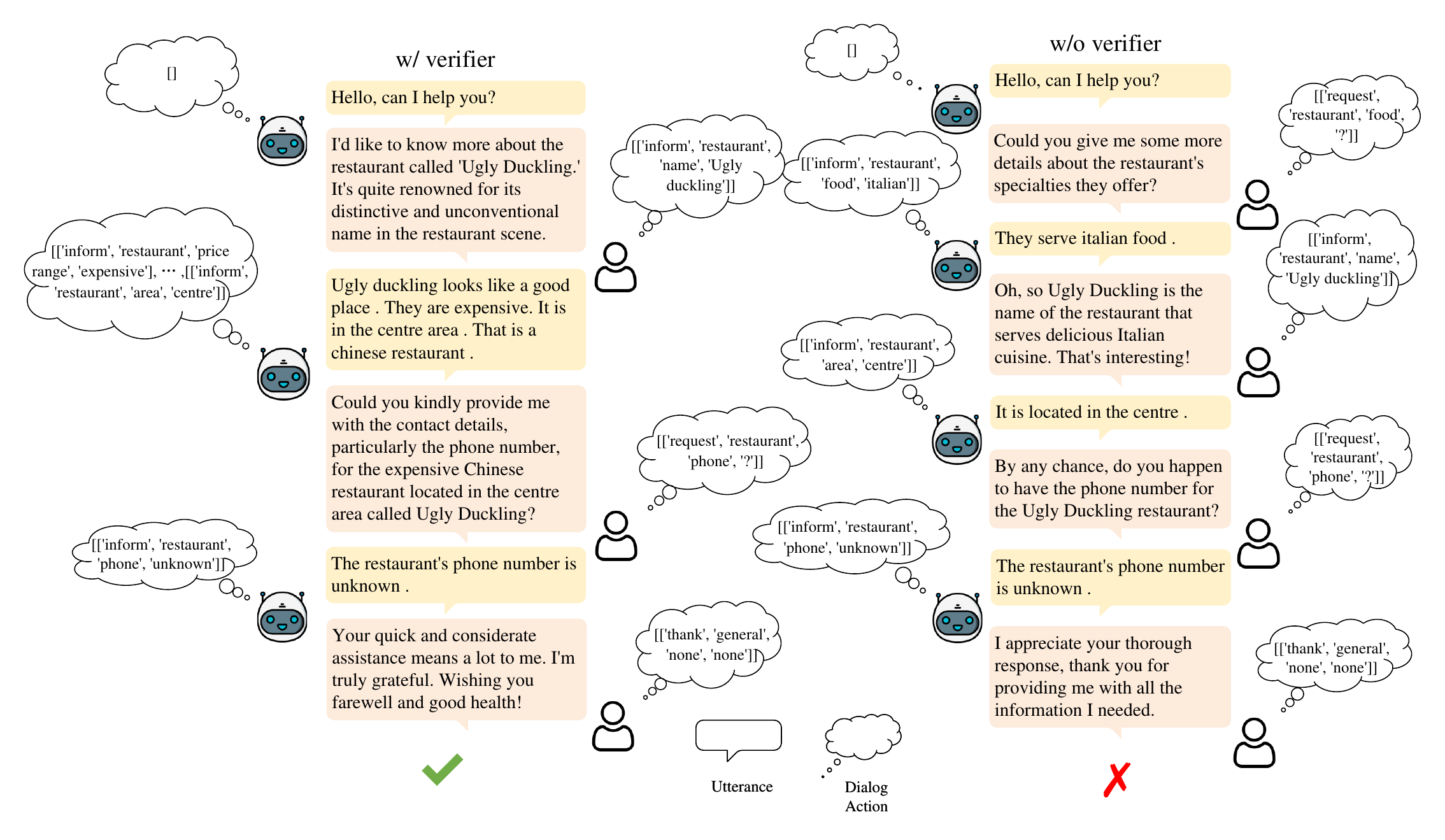}
    \caption{Dialogues generated by the model w/ and w/o verifier.}
    \label{fig:case study}
\end{figure*}

\subsection{Human Evaluation}
Apart from evaluating the user simulator with automatic metrics, we further conduct human evaluation to study human user's preference towards different user simulators. The human evaluation involves four user simulators, ABUS-T, ABUS-S, DuetSim (ChatGPT), DuetSim (FLAN-T5).

We recruited 20 human annotators to take part in the experiments.  For each user simulator, we first generate 50 task-oriented dialogues by letting the user simulator interact with a rule-based dialogue system. In total, we generated 200 dialogues for human evaluation. And we evaluated dialogues from three dimensions, namely naturalness, informativeness, and coherence. We ask annotators to rate each dialogue among 0 (poor), 1 (average),  and 2(good) for each dimension. 

Table \ref{human evaluation} reports the average score for the experiment.  We found that DuetSim (ChatGPT) performs better on naturalness and informativeness compared with all other user simulators. DuetSim (ChatGPT) also performs on par with ABUS-T in terms of coherence. Such results show that DuetSim is able to generate response that better fit human user's preference. 

In the meantime, we observe that DuetSim (FLAN-T5) does not perform well in human evaluation, though it is one of the top performer when evaluating with automatic metrics. When inspecting the output of FLAN-T5, we found that FLAN-T5 often fail to generate proper utterance from dialogue acts. Such results show that the discrepancies between automatic metrics and human evaluation for task-oriented dialogues cannot be neglected. 

\begin{table}[!t]
\resizebox{0.49\textwidth}{!}{
\begin{tabular}{llll}
\hline
User Simulator& Naturalness & Informativeness & Coherence \\ \hline
ABUS-T   & 1.40        & 1.08            & \textbf{1.38}      \\
ABUS-S   & 1.44        & 1.32            & 1.22      \\
\makecell[l]{DuetSim \\~~(ChatGPT)}  & \textbf{1.60}        & \textbf{1.42}            & 1.36      \\
\makecell[l]{DuetSim \\~~(FLAN-T5)}  & 0.66        & 0.30            & 0.52      \\ \hline
\end{tabular}
}
\caption{Human preference evaluation on MultiWOZ.}
\label{human evaluation}
\end{table}

\subsection{Case Study}
Fig \ref{fig:case study} presents the dialogue of different models when goal is set as follows:  \textit{"You are looking forward to trying local restaurants. You are looking for a particular restaurant. Its name is called ugly duckling. Once you find a restaurant, make sure you get phone number. Make sure to ask about what food it serves."}
User simulators must meet goal requirements. However, models without a response verifier may make direct food inquiries from a restaurant without specifying the restaurant's name for their intended dining location. This lack of restaurant information prevents the system from offering the correct details, leading to inaccurate responses, such as identifying the restaurant as Italian when the user wants to reserve a table at a Chinese restaurant. Consequently, a dialogue system lacking restaurant information fails to align with goal requirements, resulting in an unsuccessful dialogue. In contrast, DuetSim, equipped with a response verifier, identifies errors and provides feedback, enabling the generator to focus on specific prompts. As a result, our model can accurately deliver semantic actions, ensuring the system provides the correct information.

\section{Conclusion}
In conclusion, this paper introduces a zero-shot user simulator that is based on dual large language models, which consists of a generator and a verifier. The generator first generates  draft response while the verifier examines the draft response and provides feedback. We achieve this by in-context learning and employing chain of thoughts to produce natural and high-quality natural language responses. Empirical experiments show that our model get competitive results on MultiWOZ. 

Future work will focus on extending methods to the multi-modal task-oriented dialogue domain or attempting to address issues with large language models not paying attention to intermediate portions in long-context tasks.



\section*{Acknowledgements}
This work was supported by the National Natural Science Foundation of China (NSFC) under Grant Nos. 61966038 and 62266051, the Yunnan University Talent Research Startup Support Project under grant No. CY22623101 and No. CZ22623101, the Postgraduate Research and Innovation Foundation of Yunnan University under Grant No. KC-23234274 and the Exam-Exempted Postgraduate Research and Innovation Foundation of Yunnan University under Grant No. TM-23236972. The authors would like to thank the anonymous reviewers for their constructive comments.

\section*{Bibliographical References}

\bibliographystyle{lrec-coling2024-natbib}
\bibliography{lrec-coling2024-example}


\end{document}